\documentclass[sigconf]{acmart}
\usepackage{tabularx}
\usepackage{makecell}

\AtBeginDocument{%
  \providecommand\BibTeX{{%
    \normalfont B\kern-0.5em{\scshape i\kern-0.25em b}\kern-0.8em\TeX}}}

\setcopyright{acmcopyright}
\copyrightyear{}
\acmYear{}
\acmDOI{}
\acmDOI{}

\acmConference[CODS-COMAD '21]{8th ACM IKDD CODS and 26th COMAD }{Jan 02--04, 2021}{IIIT Bangalore, India}
\acmBooktitle{8th ACM IKDD CODS and 26th COMAD, Jan 02--04, 2021, IIIT Bangalore, India}
\acmPrice{}
\acmISBN{}

\begin{document}

\title{Neural Compound-Word (Sandhi) Generation and Splitting in Sanskrit Language}


\author{Sushant Dave}
\affiliation{\institution{Indian Institute of Technology - Delhi}}
\email{sushant.dave@gmail.com}

\author{Arun Kumar Singh}
\affiliation{\institution{Indian Institute of Technology - Delhi}}
\email{mailtoarunkus@gmail.com}

\author{Dr. Prathosh A. P.}
\affiliation{\institution{Indian Institute of Technology - Delhi}}
\email{prathoshap@ee.iitd.ac.in}

\author{Prof. Brejesh Lall}
\affiliation{\institution{Indian Institute of Technology - Delhi}}
\email{brejesh@ee.iitd.ac.in}


\begin{abstract}
This paper describes neural network based approaches to the process of the formation and splitting of word-compounding, respectively known as the Sandhi  and Vichchhed, in Sanskrit language. Sandhi is an important idea essential to morphological analysis of Sanskrit texts. Sandhi leads to word transformations at word boundaries. The rules of Sandhi formation are well defined but complex, sometimes optional and in some cases, require knowledge about the nature of the words being compounded. Sandhi split or Vichchhed is an even more difficult task given its non uniqueness and context dependence. In this work, we propose the route of formulating the problem as a sequence to sequence prediction task, using modern deep learning techniques. Being the first fully data driven technique, we demonstrate that our model has an accuracy  better than the existing methods on multiple standard datasets, despite not using any additional lexical or morphological resources. The code is being made available at https://github.com/IITD-DataScience/Sandhi\_Prakarana
\end{abstract}

\ccsdesc[500]{Computing methodologies~Learning paradigms}

\keywords{Bi-LSTM, Sanskrit, Word Splitting, Compound Word, Morphological Analysis}

\maketitle

\section{Introduction}
\label{intro}
Sanskrit is one of the oldest of the Indo-Aryan languages. The oldest known Sanskrit texts are estimated to be dated around 1500 BCE. It is the one of the oldest surviving languages in the world. A large corpus of religious, philosophical, socio-political and scientific texts of multi cultural Indian Subcontinent are in Sanskrit. Sanskrit, in its multiple variants and dialects, was the Lingua Franca of ancient India ~\cite{Harold:90}. Therefore, Sanskrit texts are an important resource of knowledge about ancient India and its people. Earliest known Sanskrit documents are available in the form called {\em Vedic Sanskrit}. {\em Rigveda}, the oldest of the four Vedas, that are the principal religious texts of ancient India, is written in {\em Vedic Sanskrit}. In sometime around 5\textsuperscript{th} century BCE, a Sanskrit scholar named {\em pARini}~\cite{Cardona:97} wrote a treatise on Sanskrit grammar named {\em azwADyAyI}, in which {\em pARini} formalized rules on linguistics, syntax and grammar for Sanskrit. {\em azwDyAyI}\footnote{\url{https://www.britannica.com/topic/Ashtadhyayi}} is the oldest surviving text and the most comprehensive source of grammar on Sanskrit today. {\em azwADyAyI} literally means eight chapters and these eight chapters contain around 4000 sutras or rules in total. These rules completely define the Sanskrit language as it is known today. {\em azwADyAyI} is remarkable in its conciseness and contains highly systematic approach to grammar. Because of its well defined syntax and extensively well codified rules, many researchers have made attempts to codify the {\em pARini’s} sutras as computer programs to analyze Sanskrit texts.

\subsection{Introduction of Sandhi and Sandhi Split in Sanskrit}
\label{SandhiIntro}
Sandhi refers to a phonetic transformation at word boundaries, where two words are combined to form a new word. Sandhi literally means 'placing together' ({\em samdhi}-{\em sam} together + {\em daDAti} to place) is the principle of sounds coming together naturally according to certain rules codified by the grammarian {\em pARini} in his {\em azwADyAyI}. There are 3 different types of Sandhi as defined in {\em azwADyAyI}.
\begin{enumerate}
\item {\em Swara} (Vowel) Sandhi: Sandhi between 2 words where both the last character of first word and first character of second word, are vowels.
\item {\em Vyanjana} (Consonant) Sandhi: Sandhi between 2 words where at least one among the last character of first word and first character of second word, is a consonant.
\item {\em Visarga} Sandhi: Sandhi between 2 words where the last character of first word is a {Visarga}(a character in Devanagri script that when placed at end of a word, indicate an additional {\em H} sound)
\end{enumerate}
An example for each type Sandhi is shown below:
\begin{quote}
\begin{verbatim}
vidyA + AlayaH = vidyAlayaH (Vowel Sandhi)
vAk + hari = vAgGari (Consonant Sandhi)
punaH + api = punarapi (Visarga Sandhi)
\end{verbatim}
\end{quote}

Sandhi Split on the other hand, resolves Sanskrit compounds and
“phonetically merged” (sandhified) words into its constituent morphemes. Sandhi Split comes with additional challenge of not only splitting of compound word correctly, but also predicting where to split. Since Sanskrit compound word can be split in multiple ways based on multiple split locations possible, split words may be syntactically correct but semantically may not be meaningful.
\begin{quote}
\begin{verbatim}
tadupAsanIyam = tat + upAsanIyam
tadupAsanIyam = tat + up + AsanIyam
\end{verbatim}
\end{quote}

\subsection{Existing Work on Sandhi}
The current resources available for doing Sandhi in open domain are not very accurate. Three most popular publicly available set of Sandhi tools viz. JNU\footnote{\url{http://sanskrit.jnu.ac.in/sandhi/gen.jsp}}, UoH\footnote{\url{http://tdil-dc.in/san/sandhi/index_dit.html}} \& INRIA\footnote{\url{https://sanskrit.inria.fr/DICO/sandhi.fr.html}} tools are mentioned in table ~\ref{tool-table}. 

\begin{table*}
  \caption{Sandhi Tools Summary}
  \label{tool-table}
  \begin{tabular}{|p{4cm}|p{10cm}|}
    \toprule
     \bf Tool Name  & \bf  Description\\
    \midrule
JNU Sandhi Tools  & This application has been developed under the supervision of Dr. Girish Nath Jha from JawaharLal Nehru University. It facilitates Sandhi as well as Sandhi Split.\\
UoH Sandhi Tools & These tools were developed at the Department of Sanskrit Studies, University of Hyderabad under the supervision of Prof. Amba Kulkarni \\
INRIA Tools & Called as Sandhi Engine and developed under the guidance of Prof. Gerard Huet at INRIA. \\
    \bottomrule
  \end{tabular}

\end{table*}

An analysis and description of these tools is present in the paper on Sandhikosh ~\cite{bhardwaj-etal-2018-sandhikosh}. The same paper introduced a dataset for Sandhi and Sandhi Split verification and compared the performance of the tools in table ~\ref{tool-table} on that dataset. 
Neural networks have been used for Sandhi Split by many researchers, for example ~\cite{aralikatte-etal-2018-sanskrit}, ~\cite{hellwig-nehrdich-2018-sanskrit} and ~\cite{Hellwig2015UsingRN}. The task of doing Sandhi has been mainly addressed as a rule based algorithm e.g. ~\cite{Raja2014ABS}. There is no research on Sandhi using neural networks in public domain so far. This paper describes experiments with Sandhi operation using neural networks and compares results of suggested approach with the results achieved using existing Sandhi tools ~\cite{bhardwaj-etal-2018-sandhikosh}.

\subsection{Existing Work on Sandhi Split}

Many researchers like ~\cite{Huet-2005} and ~\cite{Kulkarni-2009} have tried to codify {\em pARini’s} rules for achieving Sandhi Split along with a lexical resource. ~\cite{natarajan-charniak-2011-s3} proposed a statistical method based on Dirichlet process. Finite state methods have also been used ~\cite{mittal-2010-automatic}. A graph query method has been proposed by ~\cite{krishna-etal-2016-word}.

Lately, Deep Learning based approaches are increasingly being tried for Sandhi Split. ~\cite{Hellwig2015UsingRN} used a one-layer bidirectional LSTM to two parallel character based representations of a string. ~\cite{reddy-etal-2018-building} and ~\cite{hellwig-nehrdich-2018-sanskrit} proposed deep learning models for Sandhi Split at sentence level. ~\cite{aralikatte-etal-2018-sanskrit} uses a double decoder model for compound word split.  The method proposed in this paper describes an RNN based, two stage deep learning method for Sandhi Split of isolated compound words without using any lexical resource or sentence information.

In addition to above, there exist multiple Sandhi Splitters in the open domain. The prominent ones being JNU Sandhi Splitter \footnote{\url{ http://sanskrit.jnu.ac.in/sandhi/viccheda.jsp}}
, UoH Sandhi Splitter \footnote{\url{ http://sanskrit.uohyd.ac.in/scl/}}
and INRIA Sanskrit reader companion.
The paper ~\cite{aralikatte-etal-2018-sanskrit} compares the performance of above 3 tools with their results. This was an attempt to create benchmark in the area of Sanskrit Computational Linguistics.

\section{Motivation}
Analysis of sandhi is critical to analysis of Sanskrit text. Researchers have pointed out how a good Sandhi Split tool is necessary for a good coverage morphological analyzer ~\cite{Bharati2006BuildingAW}. Good Sandhi and Sandhi Split tools facilitate the work in below areas.
\begin{itemize}
\itemsep0em 
\item Text to speech synthesis system
\item Neural Machine Translation from non-Sanskrit to Sanskrit language and vice versa
\item Sanskrit Language morphological analysis
\end{itemize}

Most Sandhi rules combine phoneme at the end of a word with phoneme at the beginning of another word to make one or two new phonemes. It is important to note that Sandhi rules are meant to facilitate pronunciation. The transformation affects the characters at word boundaries and the remaining characters are generally unaffected. The rules of Sandhi for all possible cases are laid out in {\em azwADyAyI} by {\em pARini} as 281 rules. The rules of Sandhi are complex and  in some cases require knowledge of the words being combined, as some rules treat different categories of words differently. This means that performing Sandhi requires some lexical resources to indicate how the rules are to be implemented for given words. Work done in Sandhikosh ~\cite{bhardwaj-etal-2018-sandhikosh} shows that currently available rule based implementations are not very accurate. This paper tries to address the problem of low accuracy of existing implementations using a machine learning approach.  


\section{Method}
\subsection{The Proposed Sandhi Method}
\label{Sandhi_Method}
Sandhi  task is similar to language translation problem where a sequence of characters or words produces another sequence. RNNs are widely used to solve such problems. Sequence to sequence model introduced by ~\cite{sutskever2014sequence}  is especially suited to such problems, therefore same was used in this work. The training and test data were in ITRANS Devanagari format \footnote{\url{https://www.aczoom.com/itrans/}}. This data was converted to SLP1 \footnote{\url{https://en.wikipedia.org/wiki/SLP1}} as SLP1 was found more suited for proposed approach. The code was implemented in python 3.5 with Keras API running on TensorFlow backend.

\begin{figure*}[h]
  \centering
  \includegraphics[width=\linewidth]{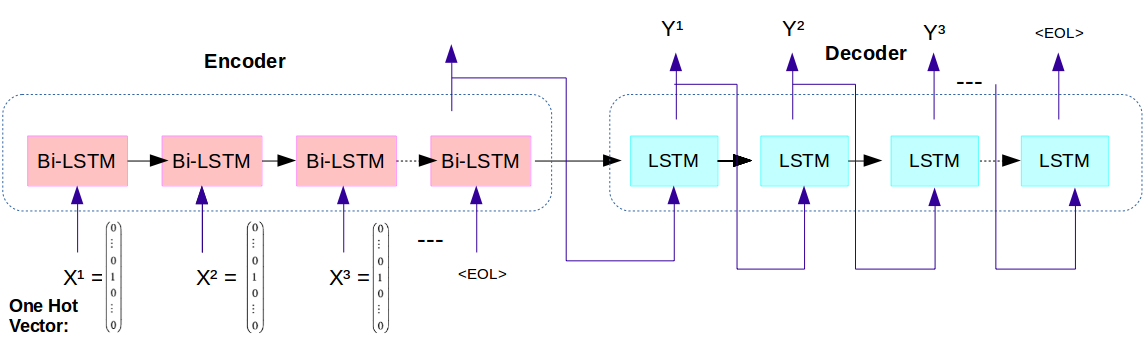}
  \caption{Model Architecture for Sandhi}
  \label{fig:sandhi-window}
\end{figure*}

Proposed model expects 2 words as input and outputs a new word as per the Sandhi rules of {\em azwADyAyI} as given in below  example.
\begin{quote}
\begin{verbatim}
sAmAnyaDvaMsAn + aNgIkArAt => sAmAnyaDvaMsAnaNgIkArAt
\end{verbatim}
\end{quote}

Results achieved from this approach are rather poor. Analysis showed that this is mainly due to the excessive length of words in some cases. Proposed approach tries to address this problem by taking last \(n\) characters of the first input word and first \(m\) characters of the second input word to do Sandhi between the 2 smaller new words. Resulting compound word is appended with the characters which are omitted from first and second input word at the beginning and end of the compound word, respectively. This approach works well for long words but suffers from the problem of losing information after truncation of words. Some Sandhi rules are specific to word category and truncated words lose the category information. Proposed method does not use any external lexical resource and learns to incorporate the word category related rules if possible and therefore words should be truncated without losing too much information.

It was found that best results were achieved with \(n=5\) and \(m=2\) for Seq2seq model used in the paper. Below example explains the truncation approach as mentioned above:

\begin{quote}
\begin{verbatim}
sAmAnyaD   vaMsAn + aN  gIkArAt 
        => sAmAnyaD  + "vaMsAn + aN” + gIkArAt
        => sAmAnyaD  + vaMsAnaN + gIkArAt
        => sAmAnyaDvaMsAnaNgIkArAt
\end{verbatim}
\end{quote}

Input sequence is set as the two input words concatenated with a ‘+’ character between the 2 words. Output sequence is the sandhified (compound) single word. Characters ‘\&’ and ‘\$’ are used as start and end markers respectively in the output sequence. A single dictionary is created for all the characters in input and output and a unique one hot vector is assigned to each token in the dictionary. The input and output character sequences are then replaced by their one hot encoded vector sequences. The best results are achieved with LSTM ~\cite{hochreiter1997long} as basic RNN cell for decoder and bidirectional LSTM as basic RNN cell for encoder. Both the encoder and decoder use the hidden unit size as 16. The training vectors were divided in batches of 64 vectors and total of 100 epochs were run to get the final results.
\subsection{Sandhi Split Method}
\label{Sandhi_Split_Method}
Conceptually, the Sandhi model architecture explained in Section ~\ref{Sandhi_Method} can be used for Sandhi Split as well, where the input and output are swapped with compound word as input and the two initial words concatenated with  ‘+’ character as output. However, the accuracy achieved with this approach is very poor due to the reasons similar to the ones observed while doing Sandhi with full word length i.e. excessive length of words which makes training the model difficult. The solution used for Sandhi i.e. employing the method of truncation of words to train the model, is not feasible in Sandhi Split due to the multiple possibilities of split point selection in the compound word.

Other researchers have tried to solve this problem in two stages i.e. predicting the split point and then splitting the sandhified (compound) word. ~\cite{aralikatte-etal-2018-sanskrit} achieved significantly good results by suggesting a double-decoder model which operates in two stages as mentioned above. An empirical analysis of the sandhi dataset showed that the following 2 observations hold for almost all the sandhified words:

\begin{itemize}
\itemsep0em 
\item No more than {\em last 2 characters of first word} and {\em first 2 characters of second word} participate in sandhi process i.e undergo a change post sandhi.
\item The {\em last 2 characters of first word} and {\em first 2 characters of second word} combine to produce {\em no more than 5 and no less than 2 characters}.
\end{itemize}

Exceptions to above rules were found to be mostly errors and thus helped clean the dataset. Above 2 rules lead to the conclusion that the portion of compound word which needs to be analyzed for change post Sandhi, should be no more than 5 characters in length. This sequence of 5 characters, hereafter referred to as sandhi-window, becomes the target of Sandhi Split. Hence, characters before sandhi-window should belong to first word and characters after sandhi-window should belong to the second word. 
Applying this reasoning,  the method described in this paper breaks the problem of Sandhi Split in 2 stages. In Stage 1, sandhi-window is predicted using an RNN model. In Stage 2, sandhi-window predicted in stage 1 is split into 2 different words using a seq2seq model similar to the one used in sandhi described in section ~\ref{Sandhi_Method}.

\subsubsection{Sandhi Split - Stage 1} 
The task of Stage 1 is to predict the sandhi-window. The input sequence is the compound word and the target output is an integer array with same number of elements as the characters in compound word. All elements in this array are 0 except the elements corresponding to the sandhi-window characters which are set to 1. For example in Fig ~\ref{fig:sandhi-window}, the sandhi-window is considered from 13\textsuperscript{th} character to 17\textsuperscript{th} character, both inclusive.

\begin{figure*}[h]
  \centering
  \includegraphics[width=\linewidth]{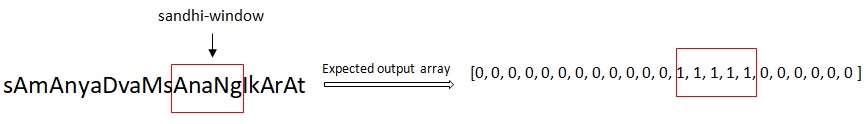}
  \caption{sandhi-window as prediction target in compound word}
  \label{fig:sandhi-window}
\end{figure*}

\noindent For an input compound word, the RNN model trained thus is expected to produce an array that has the size equal to input compound word length. All elements in this array must be zero except the elements at sandhi-window location, which must be equal to 1. For this decoder, an RNN was used with bidirectional LSTM as basic RNN cell since the LSTM cells provided slightly better overall results as compared to standard RNN cells. The output vector at each time step is connected to a dense layer with output array of unit length. This single output value is the output array element corresponding to input character. One-hot encoded vectors of input character sequences were used. Hidden unit size of Bi-LSTM cell was chosen as 64. The training vectors were divided into batch size of 64. The training was done for 40 epochs. Model was trained with RMSProp optimizer and mean squared error loss. This model when applied to a compound Sanskrit word, will produce a sequence of real numbers between 0 and 1 with size of sequence being equal to number of characters in the word. The sandhi-window is estimated to be the contiguous sequence of 5 characters that has the highest sum total for its corresponding output sequence.

\begin{figure*}[h]
  \centering
  \includegraphics[width=\linewidth]{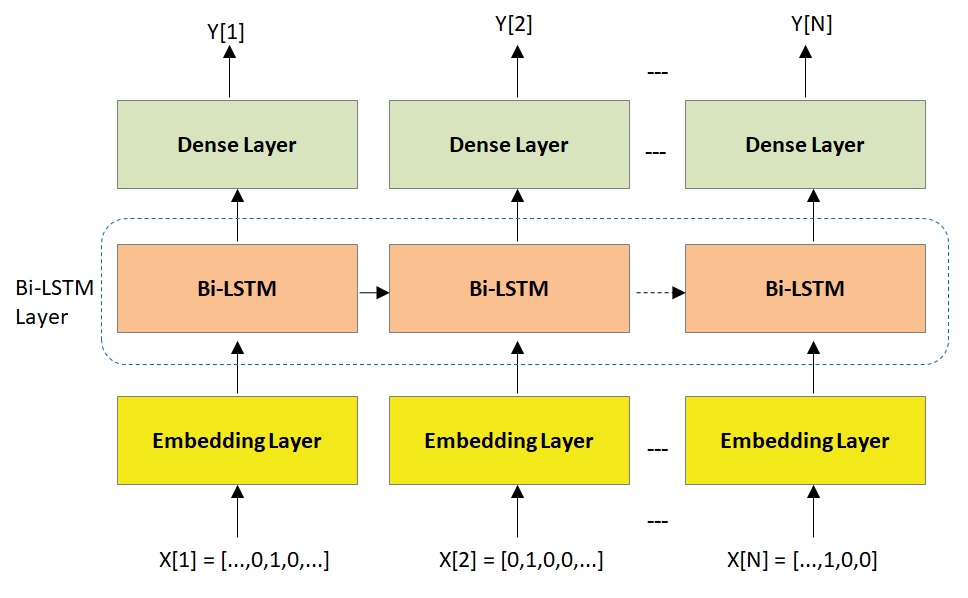}
  \caption{Model Architecture for Sandhi Split - Stage 1}
  \label{fig:model_sandhi_split}
\end{figure*}

\subsubsection{Sandhi Split - Stage 2} The task of Stage 2 is to split the sandhi-window. The model for Stage 2 uses the same architecture as the one used for Sandhi (see section ~\ref{Sandhi_Method}). Output sequence is set as truncated words \({T_{W_1}}\) and \({T_{W_2}}\) (see section ~\ref{Sandhi_Split_Evaluation} for detail) concatenated with a ‘+’ character between the 2 words. Input sequence was the sandhi-window of the compound word. Characters ‘\&’ and ‘\$’ were used as start and end markers respectively in the output sequence. A single dictionary was created for all the characters in input and output and a unique one hot vector was assigned to each token in the dictionary.The input and output character sequences were then replaced by their one hot encoded vector sequences.The best results were achieved with LSTM ~\cite{hochreiter1997long} as basic RNN cell for decoder and Bi-LSTM as basic RNN cell for encoder. Both the encoder and decoder used the hidden unit size as 128. Batch size for training was set as 64 vectors and total 30 epochs were run to get the final results.  Model was trained with RMSProp optimizer and categorical cross-entropy error loss. The detailed steps for Sandhi Split method are as follows:

\begin{itemize}
\itemsep0em 
\item Stage 1 - Predict the sandhi-window in a compound word that is to be split, using a RNN model.
\item Stage 2 - The sandhi-window is then split in 2 words using a seq2seq model. Lets call the first split of sandhi-window as \({P_{S_1}}\) and second split of sandhi-window as \({P_{S_2}}\).
\item Post-Processing step - The characters \({N_{W_1}}\) before the sandhi-window are the preceding part of first predicted split \({P_{W_1}}\) of compound word. The characters \({N_{W_2}}\) after the sandhi-window are the succeeding part of second predicted split \({P_{W_2}}\) of compound word.  \({P_{W_1}}\) is obtained by concatenating \({N_{W_1}}\) and \({P_{S_1}}\) as preceding and succeeding words respectively. \({P_{W_2}}\) is obtained by concatenating \({P_{S_2}}\) and \({N_{W_2}}\) as preceding and succeeding words respectively.
\end{itemize}

\section{Data and Evaluation Results}
\subsection{Sandhi}
\label{Sandhi_Evaluation}
The data for training a neural network model was taken from UoH corpus created at the University of Hyderabad \footnote{\url{ http://sanskrit.uohyd.ac.in/Corpus/}}. This dataset has more than 100,000 Sandhi examples. There are some errors in this dataset, some of which were removed using manually created check rules. This dataset has examples from all the 3 types of Sandhi mentioned in Section \ref{SandhiIntro}. The proposed method performs equally well on all three types of Sandhi. For current implementation, the cases where the sandhi operation is not allowed or more than 2 words combine to produce 1 or more words were discarded. Analysis showed that most Sandhi examples in our dataset followed the relationship given below.
\[-2 <= N_{c} - (N_{w_1} + N_{w_2}) <= 1 \] 
Where \({N_c}\), \({N_{w_1}}\) and \({N_{w_2}}\) are the number of SLP1 characters in compound word, first input word and second input word respectively. Of all the cases which violated this rule, most were obvious errors in dataset and the remaining cases were too few in number and discarded as outliers. It is to be noted that the above equation is consistent with the second of the two empirical rules introduced in section ~\ref{Sandhi_Split_Method}. Using this rule, total examples left in our dataset were 81029. 20\% examples out of these i.e. 16206 were separated as test set. Out of the remaining 65124 examples, 80\% examples (51858) were used for training and 20\% examples (12965) were used for model validation.
Evaluation is based on exact matching of whole compound word. Even if the model does Sandhi over the word boundaries correctly but makes an error in a character before or after, it is considered a failure.

 Results from method described above were compared with results from other publicly available tools as provided in Sandhikosha ~\cite{bhardwaj-etal-2018-sandhikosh}. Sandhikosha divides its data in 4 main sources: Ashtadhyayi, Bhagavad Gita, UoH corpus and other literature. In case of UoH corpus, test-set was selected which comes from UoH corpus and which is not used in training the seq2seq model described in this paper. For the the other 3 sets, only those test cases were chosen which have 2 input words and produce 1 output word just like it was done for training set. Since test cases used in this paper and Sandhikosha test cases are not exactly same, the results below are indicative in nature, but they do point to a clear trend.
 
The comparison is shown in the table ~\ref{table: Result_for_Sandhi}. Every cell in the table indicates successful test cases, overall test cases and success percentage.

\begin{table*}
  \caption{Benchmark Results for Sandhi}
  \label{table: Result_for_Sandhi}
  \begin{tabular}{|l|c|c|c|c|}
    \toprule
     \bf Corpus & \bf JNU  & \bf UoH & \bf INRIA & \bf Proposed Method\\
    \midrule
Literature & \makecell{53/150 \\ (14.8\%)} & \makecell{130/150 \\ (86.7\%)} & \makecell{ 128/150 \\
(85.3\%)} &  \makecell{109/115 \\ (94.78\%) }\\ & & & & \\
Bhagavad-Gita & \makecell{338/1430 \\ (23.64\%)} & \makecell{ 1045/1430 \\ (73.1\%)} & \makecell{1184/1430 \\ (82.1\%)} & \makecell{575/753 \\ (76.36\%)} \\& & & &  \\
UoH & \makecell{ 3506/9368 \\ (37.4\%)} & \makecell{7480/9368 \\ (79.8\%)} & \makecell{ 7655/9368  \\ (81.7\%)} & \makecell{ 14734/16206 \\ (90.92\%) }\\ & & & & \\
Ashtadhyayi & \makecell{ 455/2700 \\ (16.9\%)} & \makecell{1752/2700 \\ (64.9\%)} & \makecell{1762/2700 \\  (65.2\%)} & \makecell{1070/1574 \\ (68.0\%) }\\
    \bottomrule
  \end{tabular}
\end{table*}

As can be seen in the table ~\ref{table: Result_for_Sandhi}, proposed method outperforms the existing methods of doing Sandhi in every case except the INRIA Sandhi tool in case of {\em Bhagavada Gita} word-set and it does so without using any additional lexical resource.

\subsection{Sandhi Split}
\label{Sandhi_Split_Evaluation}
\begin{table*}
  \caption{Benchmark Results for Sandhi Split}
  \label{table: Results_for_Sandhi_Split}
  \begin{tabular}{|l|c|c|}
    \toprule
     \bf Model & \bf Location Prediction Accuracy & \bf Split Prediction Accuracy\\
    \midrule
        JNU &  - & 8.1\% \\
        UoH &  - & 47.2\% \\
        INRIA &  - & 59.9\% \\
        DD-RNN &  95.0\% & 79.5\% \\
        Proposed Method &  92.3\% & 86.8\% \\
    \bottomrule
  \end{tabular}
\end{table*}

The UoH corpus was used for Sandhi Split task. The same dataset was used in sandhi task also (refer section ~\ref{Sandhi_Evaluation}). Similar to approach taken for Sandhi data preparation, this dataset was converted from Devanagari format to SLP1 format. Only those examples were selected for benchmark where two words combine to produce one word. Examples which violated the two rules mentioned in section  ~\ref{Sandhi_Split_Method} were discarded. Of the total 77842 remaining examples, 20\% examples (15569) were used for testing and 80\% examples (62273) were used for model training. The steps for dataset preparation are as follows:

\begin{enumerate}
\itemsep0em 
\item Take an example from dataset consisting of a compound word \({C_W}\), first word \({W_1}\) and second word \({W_2}\) where \({W_1}\) and \({W_2}\) are the words which combine to produce \({C_W}\)
\item Mark the sandhi-window \({S_W}\) in \({C_W}\)
\item Let \({n_1}\) and \({n_2}\) be the number of characters in \({C_W}\) before and after \({S_W}\) respectively
\item Remove the first \({n_1}\) characters from \({W_1}\). Call the resulting word \({T_{W_1}}\)
\item Remove the last \({n_2}\) characters from \({W_2}\). Call the resulting word \({T_{W_2}}\)
\item The compound word \({C_W}\) and the location of sandhi-window pair is the data example for Stage 1
\item The sandhi-window \({S_W}\), \({T_{W_1}}\) and \({T_{W_2}}\) tuple makes data example for Stage 2
\item Repeat the above step for all the example selected from UoH Corpus for Sandhi-split
\end{enumerate}

The results thus obtained for the 15569 test examples, were used for benchmarking. Criteria of correct prediction was based on exact word match between actual split words and predicted split words. We compared our test results with seq2seq paper ~\cite{aralikatte-etal-2018-sanskrit} in table ~\ref{table: Results_for_Sandhi_Split}, on dataset taken from same source. The number of examples in full dataset is also sufficiently close (77842 for us vs. 71747 for ~\cite{aralikatte-etal-2018-sanskrit})
The table shows the results provided by  (~\cite{aralikatte-etal-2018-sanskrit}, Table 1) along with results achieved with approach described in this paper ( mentioned in last row of the table). To evaluate the results for JNU, UoH and INRIA tool, Sandhi Split ground truth was matched with top 10 results returned by these tools and if any of the results matched, it was considered a success.

It is evident from the results that proposed method improves upon the existing state of the art methods by a decent margin. In addition, proposed models are much simpler and do not require attention mechanism thereby reducing model complexity as well as training and inference time.

\section{Conclusion}
In this research work, we propose novel algorithms for Sandhi word formation and Sandhi Split that can be trained without use of any external resources such as language models, morphological or phonetic analyzers, and still manage to match or outperform existing approaches. Due to the simplicity of the models, these are computationally inexpensive to train and execute. In future we intend to extend current work to internal Sandhi and internal Sandhi-split using machine learning methods. Proposed models can be further refined by using additional training data as well as investigating techniques to reduce the errors in current training data.

\bibliographystyle{ACM-Reference-Format}
\bibliography{sandhi}

\end{document}